\documentclass{article}

\usepackage{graphicx}
\graphicspath{{./images/}}
\usepackage{amssymb}

\usepackage{subfig}
\usepackage{multirow}
\usepackage[section]{placeins}

\usepackage{authblk}

\providecommand{\keywords}[1]{\textbf{\textit{Index terms---}} #1.}

\begin{document}
	
\title{On the use of convolutional neural networks for robust classification of multiple fingerprint captures}

\date{\vspace{-5ex}}

\author[1,2]{Daniel~Peralta}
\author[3]{Isaac~Triguero}
\author[4]{Salvador~Garc\'ia}
\author[1,2]{Yvan~Saeys}
\author[4]{Jose~M.~Benitez}
\author[4,5]{Francisco~Herrera}

\affil[1]{\small{Department of Applied Mathematics, Computer Science and Statistics, Ghent University, Ghent, Belgium}}
\affil[2]{Data Mining and Modelling for Biomedicine group, VIB Center for Inflammation Research, Ghent, Belgium}
\affil[3]{School of Computer Science, University of Nottingham, Jubilee Campus,
	Wollaton Road, Nottingham NG8 1BB, United Kingdom}
\affil[4]{Department of Computer Science and Artificial Intelligence of the University of Granada, 18071 Granada, Spain}
\affil[5]{Faculty of Computing and Information Technology, King Abdulaziz University, Jeddah, Saudi Arabia}
\affil[ ]{e-mails: daniel.peralta@irc.vib-ugent.be, Isaac.Triguero@nottingham.ac.uk, salvagl@decsai.ugr.es, yvan.saeys@ugent.be, J.M.Benitez@decsai.ugr.es, herrera@decsai.ugr.es}

\maketitle

\begin{abstract}
	
	Fingerprint classification is one of the most common approaches to accelerate the identification in large databases of fingerprints. Fingerprints are grouped into disjoint classes, so that an input fingerprint is compared only with those belonging to the predicted class, reducing the penetration rate of the search. The classification procedure usually starts by the extraction of features from the fingerprint image, frequently based on visual characteristics.
	In this work, we propose an approach to fingerprint classification using convolutional neural networks, which avoid the necessity of an explicit feature extraction process by incorporating the image processing within the training of the classifier. Furthermore, such an approach is able to predict a class even for low-quality fingerprints that are rejected by commonly used algorithms, such as FingerCode. The study gives special importance to the robustness of the classification for different impressions of the same fingerprint, aiming to minimize the penetration in the database.
	In our experiments, convolutional neural networks yielded better accuracy and penetration rate than state-of-the-art classifiers based on explicit feature extraction. The tested networks also improved on the runtime, as a result of the joint optimization of both feature extraction and classification.
	
	\keywords{Convolutional neural networks (CNNs), fingerprint classification, deep learning, deep neural networks (DNNs)}
\end{abstract}

\section{Introduction}	\label{sec:intro}

Fingerprint identification has become the most widespread manner to implement biometric authentication~\cite{Maltoni2009}, by virtue of the desirable properties of fingerprints including uniqueness, universality and invariability. Given a database of template fingerprints, the identification consists of finding the template that corresponds to the identity of an input fingerprint.
Multiple fingerprint matching and identification algorithms have been published along the last two decades~\cite{Ratha2000,Jiang2000,Cappelli2010}. Each of them has different properties, which in turn yield different trade-offs between efficiency and accuracy~\cite{Peralta2015}. However, in a basic identification framework the input fingerprint must be compared with every template in the database, a procedure that becomes prohibitively time-consuming when dealing with extremely large databases. Therefore, it is necessary to combine these approaches with additional processing steps aimed at reducing the so-called database penetration rate.

Fingerprint classification is one of the most popular ways to achieve this goal~\cite{Galar2015}. Several classes of fingerprints are established, and the input fingerprint is classified prior its identification. Then, it is compared only to the templates belonging to the predicted class~\cite{Maltoni2009}.

Traditionally, an expert manually labels every template fingerprint in the database. Then, the classifier is trained on the obtained labeled dataset, with the aim of assigning to each input fingerprint the same label that was manually established for the corresponding template. This is a laborious and human-dependent process. Therefore, although in this paper we still consider this evaluation procedure to allow for comparison with other works on the topic, we also focus on the classification \textit{robustness}, which we define as the capacity of assigning the same class to different impressions of the same fingerprint, independently of the manual label. This enables the possibility of further increasing the performance for fingerprints that fall close to the frontier between classes.

The overall fingerprint classification process is composed of two main steps~\cite{Galar2015}: feature extraction and the classification itself. First, the captured image of the fingerprint is processed to extract meaningful features that can lead to a high discernibility between the classes. These features are frequently represented in the form of a numeric vector~\cite{Galar2015}. Second, the feature vector is used to perform the classification, either by a set of fixed rules or by training a model in a supervised manner.

Various fingerprint classification approaches have been proposed so far, based on different features such as orientation maps~\cite{Candela1995}, singular points~\cite{Hong2008,Liu2010}, ridge structure~\cite{Kawagoe1984,Candela1995} and filter-based response~\cite{Jain1999a,Hong2008}, and different ways to extract them. Most methods apply general purpose classification algorithms such as $k$ nearest-neighbors ($k$-NN)~\cite{Cover1967} or Support Vector Machines (SVMs)~\cite{Hearst1998}. Others provide fixed classification rules that do not require a training procedure~\cite{Zhang2004,Wang2007}.

The main benefit of this two-step structure is the possibility of using highly accurate machine learning classifiers. However, the manually designed feature extraction process focuses on specific features of the fingerprint pattern, which can lead to some information loss due to the discarding of subtler shapes. Due to the visual definition of the classes, the extraction of a numerical feature vector is a complex task for which many aspects of the image can be considered. In this context, some feature extraction algorithms reject fingerprints when the image does not comply with certain quality requirements, such as being properly centered in the image. The rejection of a fingerprint aims to increase the reliability of the classification, but also hinders the database penetration rate reduction~\cite{Maltoni2009}.

Deep neural networks (DNNs)~\cite{LeCun2015,Goodfellow2016} have attracted a lot of attention from the scientific community along the last few years due to their high capability for complex pattern recognition. They have been applied over multiple problems, such as image classification~\cite{Krizhevsky2012}, digit recognition~\cite{LeCun1998}, feature extraction~\cite{Sankaran2017} or gas recognition~\cite{Liu2015}, among others.
One of the advantages of DNNs is that the neuron layers implicitly extract the information from the raw input patterns. This allows for a generic learning process that does not depend on explicitly chosen nor previously extracted features. Another advantage is the possibility of applying the networks directly on the input images, usually by Convolutional Neural Networks (CNNs). Finally, the function computed by neural networks is defined for any input pattern, which in a fingerprint classification context allows for the elimination of the rejection rate.

All these properties highlight DNNs and CNNs as potentially promising models for fingerprint classification. Some authors have published proposals in this direction, such as a succession of an autoencoder and a neural network classifier~\cite{Kulkarni2011} or a succession of 1-layer autoencoders followed by a DNN that performs the classification~\cite{Wang2014}. However, there is a lack of a complete and systematic study over the capabilities of DNNs for the fingerprint classification problem.

In this paper, we propose to use convolutional neural networks on the fingerprint classification problem with the following aims:
\begin{itemize}
	\item To evaluate the accuracy of CNNs against that of state-of-the-art classifiers based on feature extraction.
	\item To increase the classification robustness when dealing with different impressions of the same fingerprints.
	\item To minimize the penetration rate that is expected after the application of the classifier, along with the identification time.
\end{itemize}
Several fingerprint databases with different qualities and characteristics are used in the analysis, replicating the baseline study proposed in~\cite{Galar2015b}. Some of these databases were synthetically generated with the SFinGe software with realistic parameters and different quality settings. The publicly available NIST-DB4 database is also used.

The remainder of this paper is structured as follows. Section~\ref{sec:background} presents some background knowledge on fingerprint classification and deep learning. Section~\ref{sec:main} describes the different classification approaches taken into consideration in this study. The analysis of the accuracy of the proposed neural network is carried out in Section~\ref{sec:exp_accuracy}. Section~\ref{sec:exp_robustness} describes the robustness of the classification on different fingerprint impressions and the penetration rate obtained when performing identifications.
Finally, the conclusions of the study are shown in Section~\ref{sec:conclusion}.

\section{Background}	\label{sec:background}

This section provides background information about the fingerprint classification problem (Section~\ref{sec:fp_classification}), detailing the state-of-the-art approaches that will be considered in the analysis (Section~\ref{sec:ml_fe}). Section~\ref{sec:deeplearning} describes the deep learning paradigm, and Section~\ref{sec:fp_class_dl} presents previous work on applying deep learning models to fingerprint recognition and other biometric problems.

\subsection{Fingerprint classification}	\label{sec:fp_classification}

\begin{figure}[!htb]
        \centering
        \captionsetup[subfloat]{labelformat=empty,justification=centering}
        \subfloat[][Arch\par (3.7\%)]{\includegraphics[width=0.15\textwidth]{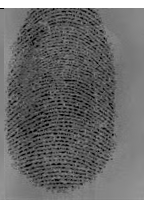}}\quad
        \subfloat[][Left Loop\par (33.8\%)]{\includegraphics[width=0.15\textwidth]{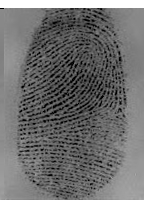}}\quad
        \subfloat[][Right Loop\par (31.7\%)]{\includegraphics[width=0.15\textwidth]{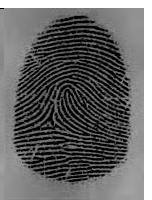}}\quad
        \subfloat[][Tented Arch\par (2.9\%)]{\includegraphics[width=0.15\textwidth]{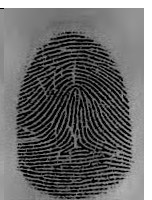}}\quad
        \subfloat[][Whorl\par (27.9\%)]{\includegraphics[width=0.15\textwidth]{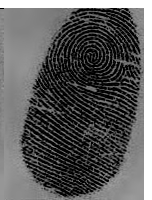}}
        \caption{Five fingerprint classes defined by Henry~\cite{Henry1900} and their frequencies.}
        \label{fig:classes}
\end{figure}

Fingerprint classification is the most common approach to reduce the database penetration rate of a fingerprint identification system~\cite{Galar2015}. The five-class system proposed by Henry~\cite{Henry1900} is still applied by most authors. These classes present different visual patterns and are unequally distributed, as shown in Fig.~\ref{fig:classes}.

\begin{figure}[!htb]
        \centering
        \captionsetup[subfloat]{justification=centering}
        \subfloat[][Orientation map]{\includegraphics[width=0.3\textwidth]{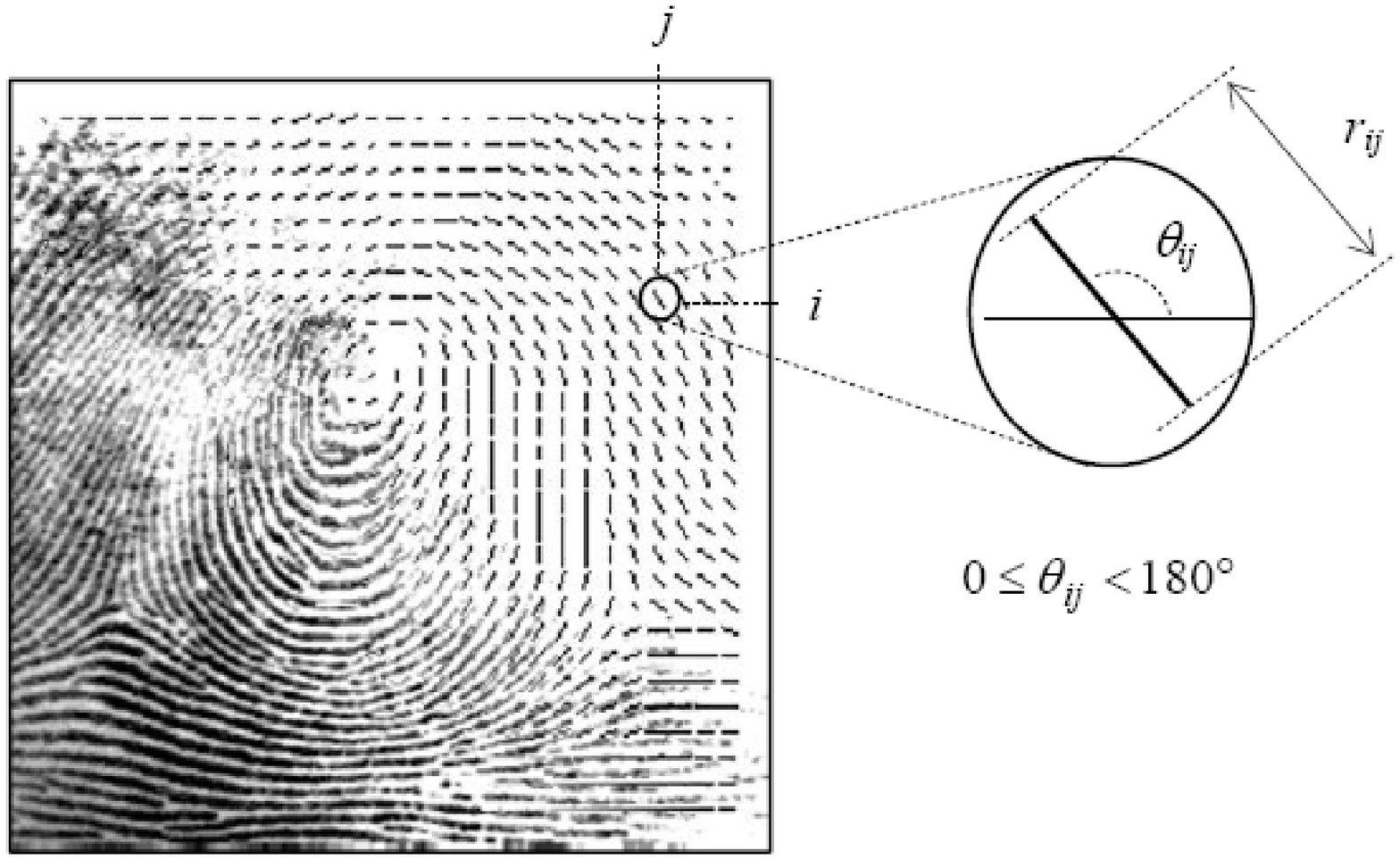}} \quad
        \subfloat[][Singular points]{\includegraphics[width=0.3\textwidth]{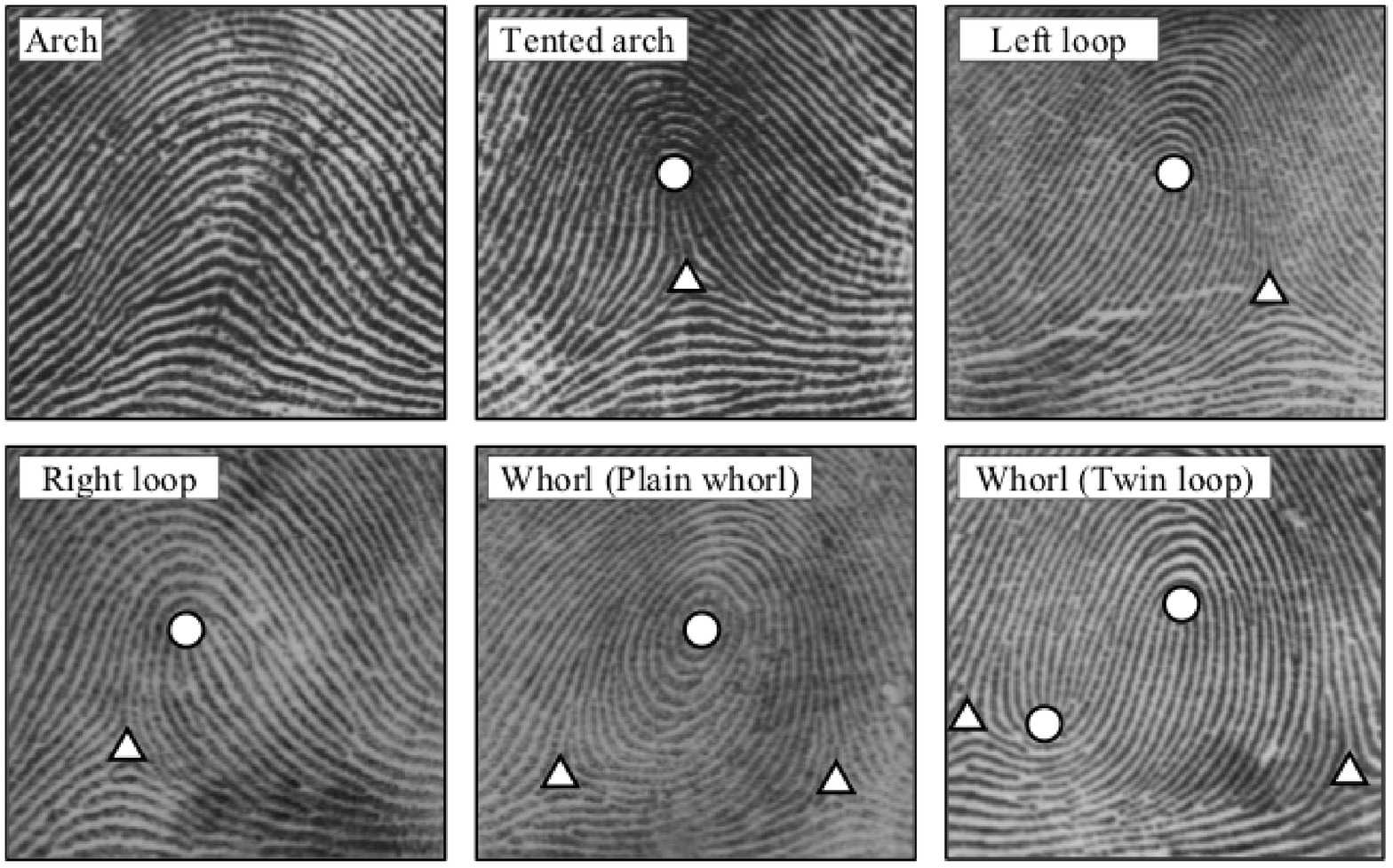}} \quad
        \subfloat[][Ridge tracing]{\includegraphics[width=0.3\textwidth]{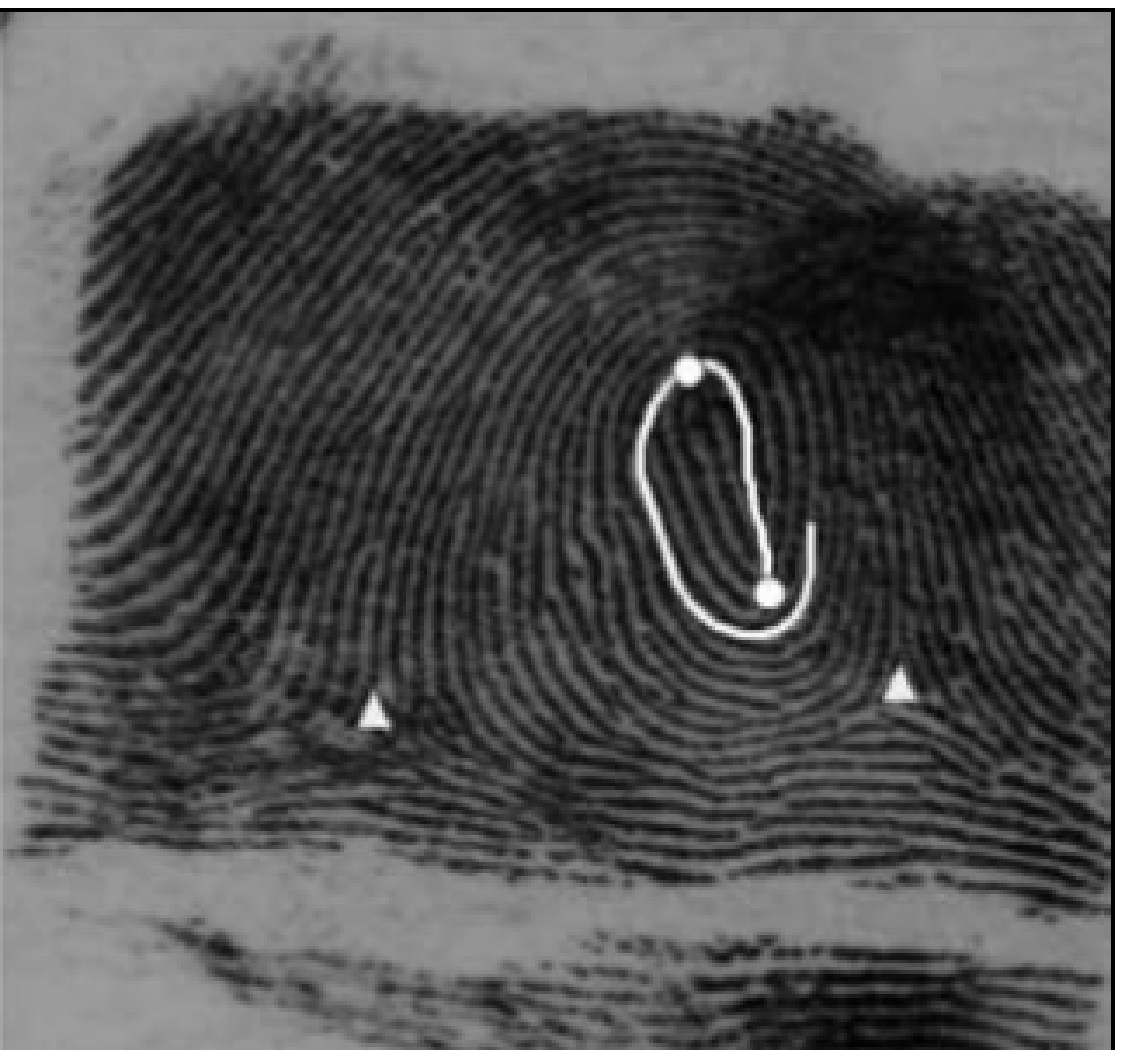}}
        \caption{Main types of global fingerprint features~\cite{Maltoni2009}.}
        \label{fig:globalfeatures}
\end{figure}

Fingerprint classification algorithms are based on the global-level features of the fingerprint images~\cite{Maltoni2009}: orientation maps, ridge structure and singular points, shown in Fig.~\ref{fig:globalfeatures}. There are many methods that extract these features, each of which provides different nuances or follows a different extraction approach. For instance, singular points are most commonly detected by the Poincar\'e method. Some feature extractors that apply this method are described in~\cite{Kawagoe1984,Karu1996,Marcialis2013}. Complex filters are also widely used for this purpose~\cite{Li2008,Liu2010,Cao2013}. Orientation maps are usually extracted by gradient-based methods~\cite{Jain1999a,Hong1998,Marcialis2013}, although there are other proposals such as slits-based~\cite{Candela1995} or skeleton tracing~\cite{Moayer1976} approaches.

Some of these extraction methods require the fingerprint to meet certain quality conditions for the extraction to be performed. This behavior ensures a meaningful feature extraction, as well as a higher classification accuracy rate for the fingerprints that are not rejected~\cite{Maltoni2009}. However, in an automatic fingerprint identification system the rejection can lead to a performance loss, as the reduction of the search space cannot be performed for fingerprints whose features cannot be extracted. Therefore, it is important to find extraction methods that lead to a high accuracy, while minimizing or eliminating the rejection rate.

Once the features have been extracted, they are used to carry out the classification itself. In many cases, the features are encoded into a numeric vector so that general purpose classifiers, such as SVMs~\cite{Min2006,Li2011}, neural networks~\cite{Candela1995,Nyongesa2004} or $k$-NN~\cite{Rajanna2010,Le2012}, can be directly trained and applied over them. Other approaches follow a more specific classification procedure. Fixed classification implements a set of fixed criteria to determine the class of a fingerprint without any training procedure~\cite{Kawagoe1984,Zhang2004,Wang2007}. Structural models rely on decision trees and hidden Markov models~\cite{Senior1997,Jung2009}. In general, however, many fingerprint classification proposals in the literature combine several of the previous systems to achieve better results.

The penetration rate of the identification search that is carried out after the classification can be estimated from the class distribution and the confusion matrix of the classifier. The estimated average penetration rate for an input fingerprint of class $i$, within a setting with $m$ classes, is shown in Eq.~\ref{eq:penetrationrateclass}, where $p_i$ is the proportion of fingerprints belonging to class $i$ and $q_i$ is the accuracy rate for that class. In the best possible scenario, $q_i=1$ and $\bar{r_i}=p_i$, that is, the classifier never misclassifies inputs of class $i$ and therefore the penetration rate for that class is always $p_i$. Eq.~\ref{eq:penetrationrate} shows the formula for the estimated penetration rate, averaged throughout all possible classes.

\begin{equation}
\label{eq:penetrationrateclass}
\bar{r_i} = 1 + q_i (1 - p_i) \quad i \in \{1, ..., m\}
\end{equation}

\begin{equation}
\label{eq:penetrationrate}
\bar{r} = \sum \limits^m_{i=1} p_i \bar{r_i}
\end{equation}

\subsection{Feature extractors and classifiers compared in the study}	\label{sec:ml_fe}

In order to meaningfully evaluate the performance of the deep learning approaches studied in this paper, several other fingerprint classification techniques from the state-of-the-art will be tested. In particular, we selected the classifiers and feature extractors that obtained the best results in~\cite{Galar2015b}, selecting algorithms with a variety of different characteristics.
Three different feature extractors have been considered, which will henceforth be referred to with the name of the first author and the year of publication. Cappelli et al.~\cite{Cappelli2002a} propose a method based on the orientation map, which is registered using the core point detected by the Poincar\'e method~\cite{Kawagoe1984}. A dynamic mask is applied for each class, producing a vector of size five. The orientations are also stored into the feature vector. Hong et al.~\cite{Hong2008} extend the FingerCode feature vector~\cite{Jain1999a} (based on Gabor filters) with the pseudo-ridges traced from the center of the fingerprint, the number of singular points (cores and deltas) and the distance and location between them. Liu's approach~\cite{Liu2010} extracts the singular points and builds a feature vector based on relative measures among them.

Three general purpose classifiers will be applied to the vectors produced by the aforementioned feature extractors. Again, we selected classifiers with very different learning procedures so as to carry out a generic study:
\begin{itemize}
\item SVM~\cite{Hearst1998}: the original feature space is mapped to a higher dimensional space by means of a kernel function, in order to make it linearly separable. The separating hyperplane is computed by maximizing the margin to the training instances in the target space.
\item Decision tree (C4.5)~\cite{Quinlan2014}: classification rules are extracted by building a decision tree from the training set, which is built in a top-down manner. At each node of the tree, the attribute with maximum difference in entropy is used to split the data. C4.5 also involves a pruning procedure.
\item $k$-NN~\cite{Cover1967}: the $k$ nearest neighbors of a test instance are computed. Then, the most frequent class among these neighbors is returned for the test instance. Therefore, the distance metric and the value of $k$ strongly determine the behavior of this classifier.
\end{itemize}

\subsection{Classification with deep neural networks}	\label{sec:deeplearning}

Neural networks have been used for decades to model all kinds of problems, due to their interesting properties, the main of which is that they are universal approximators~\cite{Hornik1989}.
Although there are multiple types of neural networks, in this paper we focus on feed-forward networks for supervised classification, as they adapt naturally to the fingerprint classification problem.
A feed-forward neural network is formed by a set of layers or neurons, each of which is connected to the neurons of the previous layer by a vector of weights, so that the value of a neuron is computed as a weighted sum of the values of the neurons in the previous layer. Additionally, neurons can apply an activation function to introduce a non-linearity. In a classification context, the instance that must be classified is used to set the values of the first layer of the network (input layer). The values are propagated along the network through one or more hidden layers until the final layer (output layer) contains the predicted class.


A DNN is a network with many hidden layers, each of which extracts---broadly speaking---a certain level of abstraction from the input pattern. Therefore, a higher number of layers allows the DNN to learn more complex and generic patterns~\cite{Nielsen2015}.
There are different types of neuron layers for DNN~\cite{LeCun1998,Nielsen2015}:
\begin{itemize}
\item Fully connected layers: each neuron is connected with weights to all the neurons in the previous layer.
\item Convolutional layers: each neuron is connected to a patch of neurons in the previous layer. The weights are shared among all the neurons of the same layer, reducing the search space of the learning process.
\item Pooling layers: usually located after a convolutional layer. As in these, each neuron is connected to a patch of the previous layer, and computes the maximum or average of those values.
\end{itemize}

In practice, networks that combine all three types of layers are called Convolutional Neural Networks (CNNs). They are well adapted to the processing of images and structures with some spatial relation, as shown by the good results obtained in different competitions~\cite{Krizhevsky2012,Russakovsky2015}.

When a network is used as a classifier for a problem with classes $c_1,...,c_m$, the output layer contains one neuron per class, forming a vector $\mathbf{a} = (a_1, ..., a_m)$. The SoftMax function (Eq.~\ref{eq:softmax}) is used to convert these values into probabilities, where $\mbox{SoftMax}(a_i)$ is the probability of the input to belong to class $c_i$. Therefore, for each instance we intend all the output neurons to produce values close to zero, except the neuron of the correct class, which should be close to one.

\begin{equation}	\label{eq:softmax}
  \mbox{SoftMax}(a_i) = \frac{e^{a_i}}{\sum\limits_{j=1}^{m} e^{a_j}} ~, ~ i = 1, ..., m
\end{equation}

The activation function used to model non-linearity is usually the Rectified Linear Unit (ReLU), which can be computed faster than the traditionally used sigmoid or hyperbolic tangent functions, and also offers interesting convergence properties~\cite{Nair2010}.

The training of a network consists of optimizing the weights of each neuron so as to obtain the desired output for each input. Therefore, the dimensionality of the search space is as high as the total number of weights. The reference algorithm for the training is back propagation with gradient descent (GD)~\cite{Rumelhart1986}.


However, GD becomes computationally expensive when applied to a DNN, due to the high dimensionality of the search space. 
Stochastic Gradient Descent (SGD) can reduce this limitation by using only a subset (or batch) of the training instances in each iteration, so that the computing of the error is biased with respect to the optimum but can be performed much faster. Each iteration over the entire training set (epoch) requires multiple iterations over the small batches. This algorithm, along with the recent advances on Graphical Processing Units (GPUs) and the availability of large datasets, has allowed for the implementation and training of DNN with very good results~\cite{Nielsen2015}.


\subsection{Fingerprint classification with deep neural networks}	\label{sec:fp_class_dl}

Despite the power that deep learning approaches offer for many classification problems, they have been scarcely applied so far to the fingerprint classification problem.

In~\cite{Kulkarni2011}, two single hidden layer perceptrons are used to classify the fingerprints into the five usual classes. The input images are cropped to 16x16 pixels, so that only a small neighborhood of the reference point of the fingerprint is taken into account. The first perceptron is an autoencoder, whose hidden layer is used as the input of the second perceptron, which performs the classification. The authors report a test accuracy of 92\%, although they do not specify the database used.
In~\cite{Wang2014}, a set of stacked 1-hidden layer autoencoders is used to learn an approximation of the identity function, so as to enhance the orientation field of the fingerprint images. Then, a 3-hidden layer neural network is applied to carry out the classification. The reported accuracy over NIST-DB4 is 93.1\%, with a 1.8\% rejection.
Other authors focus on the feature extraction step. For instance, in~\cite{Sahasrabudhe2014} good quality fingerprint images are manually selected so as to train a DNN that extracts orientations and frequencies. In~\cite{Cao2015}, the authors decompose and add noise to rolled fingerprints to train a DNN aimed to perform the orientation field estimation. A regularization of an already extracted orientation field is carried out in~\cite{Lin2016} by using autoencoders. Finally,~\cite{Schuch2016} describes a deep de-convolutional neural network to enhance the quality of fingerprint images before minutiae extraction.

Several proposals apply DNNs over the fingerprint images for the liveness detection problem~\cite{Gottschlich2016,Menotti2015,Wang2015,Kim2016}. In these proposals, the fingerprint images are divided into smaller patches that are processed independently, so as to increase the number of training examples and to simplify the processing. However, this strategy cannot be used for classification, as the class is derived from the global pattern shape of the fingerprint.

DNNs have also been applied for the recognition of other biometric characteristics, such as signature~\cite{Hafemann2016}, finger vein~\cite{Qin2015,Itqan2016} or electrocardiography~\cite{Page2015}. However, to the best of our knowledge there is no complete study of the possibilities offered by deep learning when applied to the fingerprint classification problem. This paper aims to provide a first systematic study on the field, to analyze strengths and weaknesses of DNNs in this context.

\section{Fingerprint classification strategies with deep learning}	\label{sec:main}

The study carried out on this paper involved several deep neural network architectures, described in Section~\ref{sec:dnn}. The experimental setup is detailed in Section~\ref{sec:setup}.

\subsection{Deep and convolutional neural networks}	\label{sec:dnn}
DNN architectures are usually divided into two categories: fully connected DNNs and CNNs. Although the former are in theory able to learn much more complex functions, their search space can become very large, especially when working with images, due to the enormous number of weights when the size of the neuron layers is increased. On the other hand, CNNs are very suited to work with images, limiting the computing requirements and reducing the search space of the training process of the neural network.

In the context of fingerprint classification, we have started from images fitted to a size of 227x227 pixels. This size provides sufficient quality to determine the class, without being excessively large. Therefore, the neural networks considered in the remainder of this paper involve a total of 51\,529 input neurons.

Preliminary studies carried out allowed us to discard fully connected networks for this problem, as the number of connections that have to be optimized becomes enormous even for a low number of layers. The training of such networks cannot be tackled within a single machine and would require a high-cost hardware support. Therefore, in this paper we focus on CNNs, which are much better suited to image processing.

Small CNNs such as the well-known LeNet~\cite{LeCun1998} are not powerful enough to tackle the fingerprint classification problem. This network was designed to recognize 28x28 handwritten digits and obtains a great performance on them; however, their abstraction capacity is not enough to extract the more complex patterns present in 227x227 fingerprint images.

We have considered two different CNNs for the experiments in this paper:
\begin{itemize}
\item CaffeNet: it is a variant of the famous AlexNet~\cite{Krizhevsky2012}, which obtained a very good performance on the ImageNet dataset~\cite{Deng2009}. Note that we adapted the original CaffeNet so as to better fit the image sizes and number of classes of the fingerprint classification problem. The resulting topology is shown in Table~\ref{tab:caffenet_topology}.
\item Proposal: we also developed a network with the topology shown in Table~\ref{tab:proposednet_topology}. The number of units is smaller than that of CaffeNet, which is intended at simplifying the search space of the neural network training and to accelerate the training and its convergence.
\end{itemize}

\begin{table}[htb]
\centering
\caption{Topology of the used CaffeNet variant}
\label{tab:caffenet_topology}
\resizebox{0.75\textwidth}{!}{ 
\begin{tabular}{lrrrr}\hline
\textbf{Layer type} & \textbf{Size} & \textbf{Stride} & \textbf{Grouping} & \textbf{Activation}\\ \hline
Convolutional & $11 \times 11 \times 96$ & 4 & -- & ReLU \\
Pooling & $3 \times 3$ & 2 & -- & -- \\
Convolutional & $5 \times 5 \times 256$ & 1 & 2 & ReLU \\
Pooling & $3 \times 3$ & 2 & -- & -- \\
Convolutional & $3 \times 3 \times 384$ & 1 & -- & ReLU \\
Convolutional & $3 \times 3 \times 384$ & 1 & 2 & ReLU \\
Convolutional & $3 \times 3 \times 256$ & 1 & 2 & ReLU \\
Pooling & $3 \times 3$ & 2 & -- & -- \\
Fully connected & $4096$ & -- & -- & ReLU+Dropout \\
Fully connected & $512$ & -- & -- & ReLU+Dropout \\
Fully connected & $5$ & -- & -- & SoftMax \\
\hline
\end{tabular}
}
\end{table}

\begin{table}[htb]
\centering
\caption{Topology of the proposed network}
\label{tab:proposednet_topology}
\resizebox{0.75\textwidth}{!}{ 
\begin{tabular}{lrrrr}\hline
\textbf{Layer type} & \textbf{Size} & \textbf{Stride} & \textbf{Grouping} & \textbf{Activation}\\ \hline
Convolutional & $11 \times 11 \times 48$ & 4 & -- & ReLU \\
Pooling & $3 \times 3$ & 2 & -- & -- \\
Convolutional & $5 \times 5 \times 128$ & 1 & 2 & ReLU \\
Pooling & $3 \times 3$ & 2 & -- & -- \\
Convolutional & $3 \times 3 \times 192$ & 1 & -- & ReLU \\
Convolutional & $3 \times 3 \times 128$ & 1 & 2 & ReLU \\
Pooling & $3 \times 3$ & 2 & -- & -- \\
Fully connected & $2096$ & -- & -- & ReLU+Dropout \\
Fully connected & $256$ & -- & -- & ReLU+Dropout \\
Fully connected & $5$ & -- & -- & SoftMax \\
\hline
\end{tabular}
}
\end{table}

Both networks are trained with the SGD algorithm. The input images are grayscale values between 0 and 256. The global average value of the images of the training set is first computed, and then subtracted from the images that are passed as input to the network, so that the input has a zero average, which facilitates the convergence of the learning process.

\subsection{Experimental setup}	\label{sec:setup}
The classification algorithms described in Sections~\ref{sec:ml_fe} and~\ref{sec:dnn} have been applied to five different fingerprint databases, replicating the experimental setup presented in~\cite{Galar2015b}. The parameters used for the experiments are listed in Table~\ref{tab:params}.
The publicly available KEEL software~\cite{Alcala2009} was used in the experiments for the SVM, decision trees and $k$-NN. The CNNs were implemented using the Caffe library~\cite{Jia2014}. All the experiments were carried out in a single computer with an Intel Core i7-3820 processor (3.60GHz) and 24GB RAM. The CNNs were run on a single Nvidia GeForce GTX TITAN GPU (2688 cores, 6144 MB GDDR5 RAM).

\begin{table}[!htb]
\centering
\caption{Parameters of the classifiers}

\resizebox{0.75\textwidth}{!}{ 
\begin{tabular}{ll}
\hline
\textbf{Algorithm} & \textbf{Parameters} \\ \hline

\multirow{3}{*}{\textbf{SGD (CaffeNet)}} & Batch size = 256, Iterations = 2000, \\ 
&  Learning rate = 0.01, Momentum = 0.9, $\gamma$ = 0.1,  \\
& Step size = 500, Weight decay = 0.001 \\ \hline
\multirow{4}{*}{\textbf{SGD (Proposal)}} & Batch size = 128, Learning rate = 0.01,  \\ 
& Momentum = 0.9, $\gamma$ = 0.1, Weight decay = 0.0005,  \\
& Iterations = 4000 (1300 for NIST-DB4), \\
& Step size = 1000 (220 for NIST-DB4) \\
\hline
\multirow{2}{*}{\textbf{C4.5}}  & Prune = True, Confidence level = 0.25,  \\
& Minimum number of item-sets per leaf = 2 \\ \hline
\multirow{3}{*}{\textbf{SVM}} & Kernel = Polynomial/RBF, C = 1.0, \\
 & Tolerance parameter = 0.001, $\epsilon = 10^{-12}$ \\
 & Polynomial Degree = 1, Fit Logistic Models = True \\ \hline
\textbf{$k$-NN} & Distance metric = Euclidean, $k$ = 5 \\ \hline
\end{tabular}
}
\label{tab:params}
\end{table}

These are the databases used for the study:
\begin{itemize}
\item SFinGe: in order to replicate the experiments carried out in~\cite{Galar2015b}, we used the SFinGe software~\cite{Maltoni2009,Cappelli2002} to generate three different databases of synthetic fingerprints with different qualities following the natural class distribution. This approach enables a meaningful comparison of the tested classifiers according to a common measure of quality of the fingerprints. All the fingerprints of the three databases were generated using the parameters shown in Table~\ref{tab:params_sfinge}, the only difference between them is the quality of the generated images (see Fig.~\ref{fig:sfinge}): High Quality with no Perturbations (HQNoPert), Default or Varying Quality and Perturbations (VQAndPert).
Each of the three generated databases contains four captures of 10\,000 different fingerprints.

\begin{table}[!htb]
\centering
\caption{Parameter specification used with the SFinGe tool.}
\label{tab:params_sfinge}
\resizebox{0.75\textwidth}{!}{ 
\begin{tabular}{|l|l|}\hline
\textbf{Scanner parameters} & \textbf{Generation parameters} \\ \hline
Acquisition area: 14.6mm x 19.6mm. & Impressions per finger: 4. \\
Resolution: 500 dpi. & Class distribution: Natural. \\
Image size: 288 x 384. & Save ISO templates: enabled. \\
Background type: Optical. & Generate pores: enabled.  \\
Background noise: Default.   & Output file type: WSQ. \\
Crop borders: 0 x 0. & \\
\hline
\end{tabular}
}
\end{table}

\begin{figure}[!htb]
        \centering
        \captionsetup[subfloat]{justification=centering}
        \subfloat[][HQNoPert]{\includegraphics[width=0.25\textwidth]{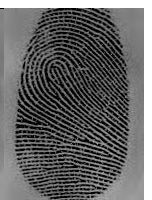}}~
        \subfloat[][Default]{\includegraphics[width=0.25\textwidth]{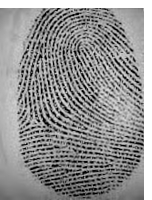}}~
        \subfloat[][VQAndPert]{\includegraphics[width=0.25\textwidth]{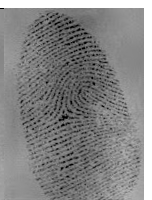}}
        \caption{Example of the quality of the three SFinGe databases.}
        \label{fig:sfinge}
\end{figure}

\item NIST-DB4: this publicly available database~\cite{Watson1992} has been extensively used by other authors to test fingerprint classification approaches. It is composed of two impressions of 2000 rolled fingerprints, manually labeled and evenly distributed among the five classes. Among them, 350 fingerprints have two labels because the visual discernibility between the classes is not absolute. Different authors treat this special case differently, so we followed the approach in~\cite{Galar2015b} and removed these fingerprints from the database. Therefore, the resulting database is composed by two impressions of 1650 fingerprints (with the class distribution shown in Table~\ref{tab:db4_distribution}), for a total of 3300 images.
\end{itemize}

\begin{table}[!htb]
\centering
\caption{Class distribution of NIST-DB4 after removing the fingerprints with two labels.}
\label{tab:db4_distribution}
\begin{tabular}{|l|r|}\hline
\textbf{Class} & \textbf{Number of fingerprints} \\ \hline
Arch   & 380 \\
Left   & 378 \\
Right  & 373 \\
Tented & 123  \\
Whorl  & 396 \\
\hline
\end{tabular}
\end{table}

\section{Analysis of the classification accuracy}	\label{sec:exp_accuracy}
This section evaluates the performance obtained with the CNNs developed in this work. The experimental study carried out in~\cite{Galar2015b} has been replicated, to enable a fair comparison between the tested CNNs and the state-of-the-art fingerprint classification methods.
The feature extractors and classifiers described in previous sections, which obtained the best results in the aforementioned study, have been applied to the described databases for a comparison of their accuracy. For this study, a single impression of each fingerprint in the SFinGe databases was used. Likewise, NIST-DB4 was split into two different databases, one for each impression. This makes a total of 5 databases for the experiments: three SFinGe databases with 10\,000 images and two NIST databases with 1650 images. The results presented in this section have been obtained with a 5-fold cross-validation scheme (5-fcv)~\cite{MorenoTorres2012}, where each database is randomly split into 5 subsets with an identical class distribution. The presented accuracy values for each database and classifier are therefore averages in test over 5 different executions.

\begin{table}[!htb]
\centering
\caption{Average cross validation test accuracy (in percentage) of different classifiers, feature extractors and ensembles. Only the best combinations are shown. In due case, the rejection rate is shown between brackets.}

\resizebox{\textwidth}{!}{
\begin{tabular}{l|rrr|rr|rr}
\hline
  & \multicolumn{3}{|c|}{\textbf{Feature extraction + classifier}} & \multicolumn{2}{c|}{\textbf{Ensembles}} & \multicolumn{2}{c}{\textbf{Deep Neural Networks}} \\ \hline
  & \multicolumn{1}{|c}{\textbf{Cappelli02}} & \multicolumn{1}{c}{\textbf{Hong08}} & \multicolumn{1}{c|}{\textbf{Liu10}} & \multicolumn{1}{c}{\textbf{HLZC-Cons}} & \multicolumn{1}{c|}{\textbf{HLC-Maj}} & \multicolumn{1}{c}{\textbf{CaffeNet}} & \multicolumn{1}{c}{\textbf{Proposed network}} \\
 & \multicolumn{1}{c}{\textbf{SVM-RBF}} & \multicolumn{1}{c}{\textbf{SVM-Poly}} & \multicolumn{1}{c|}{\textbf{C4.5}} & & & & \\ \hline

\textbf{HQNoPert} & 93.87 & 97.32 (1.44) & 94.75 & 99.47 (17.49) & 97.53 (2.27) & 98.94 & 99.07 \\ 
\textbf{Default} & 92.10 & 96.29 (5.38) & 93.96 & 99.53 (27.78) & 97.29 (6.56) & 98.06 & 98.58 \\ 
\textbf{VQAndPert} & 86.21 & 92.92 (15.90) & 90.15 & 99.10 (43.34) & 95.38 (17.89) & 97.08 & 97.54 \\ 
\textbf{NIST-DB4\_F} & 87.45 & 89.49 (1.45) & 82.67 & 98.69 (30.24) & 92.84 (4.36) & 85.09 & 90.73 \\
\textbf{NIST-DB4\_S} & 87.03 & 85.29 (1.94) & 80.61 & 98.45 (34.97) & 91.61 (5.45) & 85.52 & 88.91 \\ \hline
\end{tabular}
\label{tab:accuracy}
}
\end{table}

Table~\ref{tab:accuracy} presents the accuracy obtained for the different classification methods, split into three groups of columns. The first group shows the feature extraction methods published in the literature in combination with the classifier that obtained the best result in each case in~\cite{Galar2015b}. Note that for the sake of simplicity, only the best performing classifier is included in the table. The second group includes two ensembles presented in~\cite{Galar2015b}, which obtained the best accuracy and the best rejection, respectively. Finally, the third group includes the two deep neural networks designed in this study. The columns corresponding to Hong's feature extractor and the ensembles also show the rejection rate in brackets, that is, the proportion of fingerprints whose features could not be extracted by Hong's algorithm. The main conclusions from the table are:
\begin{itemize}
\item The accuracy obtained by the proposed network is better than that of the larger CaffeNet, despite its smaller number of layers and neurons. Although a larger network can offer better learning potential for complex problems, in this case the smaller search space provided by the proposed architecture allows for a better convergence of the learning process, yielding a very good accuracy which is far above any of those obtained by the individual classifiers with feature extraction. This result highlights the capabilities of the intrinsic feature extraction process of the deep neural network, which outperforms those manually designed independently of the subsequent classification step.
\item The proposed network also overcomes the accuracy obtained by the ensemble HLC-Maj for the SFinGe databases, despite the fact that it does not reject any fingerprint. In opposite, the ensemble HLZC-Cons, with a very large rejection rate, and HLC-Maj for NIST-DB4, are able to obtain accuracies larger than those yielded by the tested deep neural network. Nevertheless, in a practical environment it is usually preferable to eliminate such a high rejection rate at the cost of a slight reduction of the accuracy.
\end{itemize}

\section{Robust classification of different fingerprint impressions and penetration rate reduction}	\label{sec:exp_robustness}

The previous section highlighted CNNs as a powerful approach from a classic machine learning perspective. However, the biometric identification problem presents some particularities that should be taken into account when new techniques are evaluated. In practice, the template fingerprints that are stored after the enrollment of the users are grouped according to their class and used to train the classifier. When an input fingerprint is received, the goal of the fingerprint classifier is to determine the group in which the template of the same fingerprint is stored. There exists the possibility that when a template is misclassified, a corresponding input might be misclassified in the same way. In these cases, a natural way to improve the penetration is to establish the template groups not according to the manually established label of the fingerprints, but to the class that is predicted by the classifier trained upon them. Thus, if the template and input impressions of a same fingerprint are both wrongly assigned the same class, the searched would still be carried out in the group of its corresponding template. The aim of this section is to evaluate the compared classification methods from this point of view, which we refer to as \textit{robustness} of the classification. The runtime of the different methods is also discussed.

For this purpose, we used the three SFinGe databases of different qualities described in Section~\ref{sec:setup}, each of them composed by 4 impressions of the same 10\,000 fingerprints, and the NIST-DB4 database, composed by 2 impressions of 1650 fingerprints.
For each of these databases, the first impression of each fingerprint was used as training set (templates), while the remaining impressions were stripped of their manually established labels and conformed the test set (inputs). After the classifier is trained, every fingerprint was relabeled according to the class assigned by the classifier to the impression in the training set. In this manner, the accuracy reported throughout this section corresponds to the percentage of impressions that are classified into the same class as the template impression, independently of the manually established label. For the methods that reject fingerprints, only the fingerprints with no rejected impressions were considered.

\begin{table}[!htb]
\centering
\caption{Accuracy and penetration rate of the reference methods over the 4 tested databases.}
\resizebox{\textwidth}{!}{
\begin{tabular}{l|rrr|rrr|rrr|r}
\hline
& \multicolumn{3}{|c|}{\textbf{SVM-Poly}} & \multicolumn{3}{c|}{\textbf{C4.5}} & \multicolumn{3}{c|}{\textbf{KNN}} & \textbf{Rejection} \\ 
& \multicolumn{1}{c}{\textbf{Training}} & \multicolumn{1}{c}{\textbf{Test}} & \multicolumn{1}{c|}{\textbf{P. rate}} & \multicolumn{1}{c}{\textbf{Training}} & \multicolumn{1}{c}{\textbf{Test}} & \multicolumn{1}{c|}{\textbf{P. rate}} & \multicolumn{1}{c}{\textbf{Training}} & \multicolumn{1}{c}{\textbf{Test}} & \multicolumn{1}{c|}{\textbf{P. rate}} & \textbf{} \\ \hline

\textbf{Cappelli02}  & & & & & & & & & &\\
\textbf{HQNoPert} & 99.71 & 92.11 & 36.63\% & 99.27 & 94.97 & 33.85\% & 88.46 & 86.03 & 41.89\% & -- \\ 
\textbf{Default} & 98.93 & 90.38 & 38.09\% & 99.14 & 93.27 & 35.56\% & 86.79 & 82.14 & 46.78\% & -- \\ 
\textbf{VQAndPert} & 94.69 & 83.83 & 46.35\% & 98.70 & 89.80 & 40.77\% & 86.95 & 79.76 & 51.51\% & -- \\ 
\textbf{NISTDB4} & 100.00 & 86.42 & 34.61\% & 98.48 & 88.24 & \textbf{34.17\%} & 84.06 & 83.70 & 37.11\% & -- \\ \hline

\textbf{Hong08} & & & & & & & & & & \\
\textbf{HQNoPert} & 98.41 & \textbf{98.07} & 35.49\% & 99.23 & 96.19 & 37.27\% & 96.48 & 97.59 & 36.29\% & 6.71\% \\ 
\textbf{Default} & 97.67 & \textbf{96.90} & 43.97\% & 99.01 & 94.54 & 46.01\% & 96.54 & 96.58 & 44.87\% & 18.15\% \\ 
\textbf{VQAndPert} & 95.44 & 91.95 & 61.55\% & 98.48 & 91.30 & 62.38\% & 94.32 & \textbf{92.37} & 61.94\% & 40.11\% \\ 
\textbf{NISTDB4} & 98.00 & \textbf{89.65} & 35.63\% & 97.76 & 86.78 & 39.41\% & 91.08 & 87.72 & 39.49\% & 2.79\% \\ \hline

\textbf{Liu10}  & & & & & & & & & &\\
\textbf{HQNoPert} & 94.61 & 96.72 & 32.40\% & 98.06 & 95.14 & 33.75\% & 94.64 & 95.78 & \textbf{33.28\%} & -- \\ 
\textbf{Default} & 93.84 & 93.96 & \textbf{34.17\%} & 97.52 & 93.67 & 35.04\% & 94.23 & 93.48 & 34.74\% & -- \\ 
\textbf{VQAndPert} & 91.84 & 88.16 & 39.38\% & 96.53 & 90.49 & \textbf{38.05\%} & 92.08 & 88.37 & 39.03\% & -- \\ 
\textbf{NISTDB4} & 90.36 & 85.03 & 37.79\% & 93.03 & 86.55 & 37.00\% & 87.76 & 81.09 & 41.63\% & -- \\ \hline
\end{tabular}
}
\label{tab:robustness_fe}
\end{table}

The results obtained with the reference feature extractors using several classifiers are presented in Table~\ref{tab:robustness_fe}. The estimated penetration rates were calculated following Eq.~\ref{eq:penetrationrate}. The best result obtained for each database is bold-stressed. Clearly, Hong's extractor obtains a better accuracy than the rest, especially in combination with the SVM classifier, at the cost of rejecting some fingerprints. Although C4.5 can obtain a very good training accuracy, the difference with respect to the test values are larger than those obtained by other classifiers, denoting overfitting. However, the penetration rate yielded by Hong's method is higher than with other extractors at the cost of a non-negligible rejection rate.

\begin{table}[!htb]
\centering
\caption{Accuracy and penetration rate of the convolutional neural networks.}
\resizebox{0.8\textwidth}{!}{
\begin{tabular}{l|rrr|rrr}
\hline
 & \multicolumn{3}{|c|}{\textbf{CaffeNet}}  & \multicolumn{3}{c}{\textbf{Proposed network}} \\ 
 & \multicolumn{1}{|c}{\textbf{Training}} & \multicolumn{1}{c}{\textbf{Test}} & \multicolumn{1}{c|}{\textbf{P. rate}} & \multicolumn{1}{c}{\textbf{Training}} & \multicolumn{1}{c}{\textbf{Test}} & \multicolumn{1}{c}{\textbf{P. rate}} \\ \hline
\textbf{HQNoPert} & 100.00 & 99.35 & 29.99\% & 100.00 & \textbf{99.60} & \textbf{29.79\%} \\ 
\textbf{Default} & 100.00 & 97.22 & 31.59\% & 100.00 & \textbf{98.07} & \textbf{31.01\%} \\ 
\textbf{VQAndPert} & 100.00 & 95.73 & 32.77\% & 100.00 & \textbf{96.40} & \textbf{32.27\%} \\ 
\textbf{NISTDB4} & 100.00 & 90.24 & 30.89\% & 100.00 & \textbf{91.09} & \textbf{30.32\%} \\ \hline
\end{tabular}
}
\label{tab:robustness_dnn}
\end{table}

Table~\ref{tab:robustness_dnn} shows the accuracy and penetration rate obtained with the deep neural networks. Note that the training accuracy was 100\% in all cases, so the original label was kept for all fingerprints. Despite what is seemingly an extreme case of overfitting, the test results outperform those obtained with any of the combinations in Table~\ref{tab:robustness_fe}, assessing the high robustness of deep learning approaches for this problem. Moreover, these accuracy values were obtained without rejecting any fingerprint, which gives a very low penetration rate. Even though both tested networks obtain the maximum training accuracy, The proposed network performs better for the test sets due to its smaller size and the subsequent better generalization capability.

\begin{figure}[!htb]
        \centering
        \subfloat[][Arch (Left)]{\includegraphics[width=0.15\textwidth]{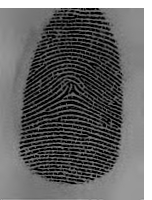}}\quad
        \subfloat[][Left (Tented)]{\includegraphics[width=0.15\textwidth]{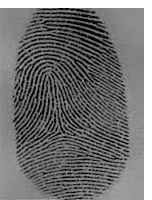}}\quad
        \subfloat[][Left (Whorl)]{\includegraphics[width=0.15\textwidth]{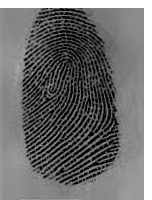}}\quad
        \subfloat[][Tented (Right)]{\includegraphics[width=0.15\textwidth]{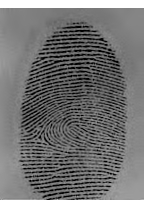}}\quad
        \subfloat[][Whorl (Left)]{\includegraphics[width=0.15\textwidth]{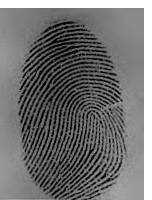}}
        \caption{Some examples of fingerprints misclassified by the proposed network. The true class is shown in brackets.}
        \label{fig:proposednet_errors}
\end{figure}

\begin{table}[htb]
\centering
\caption{Confusion matrix of the proposed network on HQNoPert}
\begin{tabular}{rrrrrr}
  \hline
 & \multicolumn{5}{c}{Predicted class} \\
  True class & A & L & R & T & W \\ 
  \hline
A & 1110 &   0 &   0 &   0 &   0 \\ 
  L &   0 & 10\,095 &   0 &  30 &  15 \\ 
  R &   0 &   0 & 9500 &  10 &   0 \\ 
  T &   0 &  40 &   0 & 830 &   0 \\ 
  W &   0 &  25 &   0 &   0 & 8345 \\ 
   \hline
\end{tabular}
\label{tab:confmatrix}
\end{table}

Note that the proposed network obtains 99.60\% accuracy for the HQNoPert database, which corresponds to only 120 failures among the 30\,000 fingerprints of the test set. The confusion matrix shown in Table~\ref{tab:confmatrix} reflects clearly the structure of the fingerprint classes: class Arch is well differentiated from the others---to the point that the network classifies correctly all 1110 test fingerprints of this class---whilst Tented is somewhat overlapped with Left and Right Loop. Some examples of failed fingerprints, along with the classes that were wrongly assigned, are shown in Figure~\ref{fig:proposednet_errors}.

\begin{table}[htbp]
\centering
\caption{Runtime (in seconds) of the feature extractors and the classifiers}

\resizebox{0.9\textwidth}{!}{
\begin{tabular}{llrrrr}
\hline
\textbf{Feature Extractor} & \textbf{Database} & \multicolumn{1}{l}{\textbf{Feature Extraction}} & \multicolumn{1}{l}{\textbf{SVM-Poly}} & \multicolumn{1}{l}{\textbf{C4.5}} & \multicolumn{1}{l}{\textbf{KNN}} \\ \hline
\textbf{Cappelli02} &\textbf{HQNoPert} & 11\,275 & 155 & 391 & 375 \\ 
& \textbf{Default} & 11\,377 & 88 & 410 & 372 \\ 
& \textbf{VQAndPert} & 11\,358 & 760 & 389 & 373 \\ 
& \textbf{NISTDB4} & 1880 & 12 & 17 & 9 \\ \hline
\textbf{Hong08} &\textbf{HQNoPert} & 3469 & 12 & 51 & 81 \\ 
& \textbf{Default} & 3381 & 16 & 66 & 103 \\ 
& \textbf{VQAndPert} & 3139 & 10 & 29 & 40 \\ 
& \textbf{NISTDB4} & 615 & 2 & 4 & 1 \\ \hline
\textbf{Liu10} &\textbf{HQNoPert} & 507 & 13 & 33 & 25 \\ 
& \textbf{Default} & 499 & 12 & 31 & 26 \\ 
& \textbf{VQAndPert} & 519 & 15 & 32 & 26 \\ 
& \textbf{NISTDB4} & 70 & 2 & 1 & 1 \\ \hline
\end{tabular}
}
\label{tab:times_fe}
\end{table}

\begin{table}[!htb]
\centering
\caption{Runtime (in seconds) of the deep neural networks.}
\resizebox{0.6\textwidth}{!}{
\begin{tabular}{lrr}
\hline
 & \multicolumn{1}{l}{\textbf{CaffeNet}} & \multicolumn{1}{l}{\textbf{Proposed network}} \\ \hline
\textbf{HQNoPert} & 2306 & 960 \\ 
\textbf{Default} & 2329 & 957 \\ 
\textbf{VQAndPert} & 2328 & 960 \\ 
\textbf{NISTDB4} & 2322 & 487 \\ \hline
\end{tabular}
}
\label{tab:times_dnn}
\end{table}

Finally, Tables~\ref{tab:times_fe} and~\ref{tab:times_dnn} show the runtime in seconds of the different feature extractors and classifiers. Liu's extractor is the fastest, followed by Hong and Cappelli. The runtime of the classifiers themselves is determined by the size of the feature vectors, so that again Liu is the fastest approach. As for the deep neural networks, the simpler proposed architecture allows for a better learning time despite the fact that it involves a higher number of iterations than CaffeNet. Note that for the proposal the number of iterations of the learning process was set lower for NIST-DB4 than for SFinGe due to the smaller size of the database; this allows to further reduce the time needed for the training of the network while maintaining a high accuracy. In general, the network is able to learn a better model than the approaches using Cappelli's or Hong's features in a lower time. Liu's extractor is however yet faster, although the accuracy gain obtained by the CNN is high enough so as to make this time consumption acceptable. Note that an ensemble using several feature extractors would consume at least the sum of their runtimes, highlighting the power of CNNs. Moreover, the rejection carried out by some feature extractors can further increase the time as a new capture of the fingerprint might be needed.

\section{Conclusions}	\label{sec:conclusion}
Fingerprint classification is a key component in automatic fingerprint identification systems that deal with very large-scale databases, as it enables the reduction of the database penetration of the identification process. A good accuracy rate for the classification is critical to maximize this reduction. Although some feature extractors and classifiers reject fingerprints to increase the accuracy rate, this rejection can also have a negative impact on the throughput of such an identification system.

In this paper, we have presented a study over the performance that can be expected from deep learning approaches when applied to the fingerprint classification problem. Several state-of-the-art fingerprint feature extractors and classifiers have been compared with two different CNN architectures, in an experimental study involving three artificially generated databases of different qualities and the well-known NIST-DB4. Also, the experiments have been focused on both the accuracy of the classification process and the robustness when dealing with several impressions of the same fingerprints.

The obtained results highlighted that DNNs outperformed all the compared approaches. The classification accuracy reached by CNNs was superior to that obtained by any of the combinations of feature extractors and classifiers. Moreover, CNNs do not reject any fingerprints, but still obtained better accuracy than feature extractors with a certain rejection rate. The robustness experiments also showed that the deep learning strategy was able to obtain a very high test accuracy, assessing that the models learned the underlying structure of the fingerprints better than state-of-the-art feature extractors and classifiers. The runtime required by CNNs was also very competitive with respect to that needed by the combination of explicit feature extraction and classification, outperforming most approaches, and in particular the most accurate ones.

\section*{Acknowledgements}
This work was partially supported by the Spanish Ministry of Science and Technology under the project TIN2014-57251-P, and the Foundation BBVA project 75/2016 BigDaPTOOLS. Y. Saeys is an ISAC Marylou Ingram Scholar.

\bibliographystyle{ieeetr}
\bibliography{biblio.bib}

\end{document}